\documentclass[letterpaper, 10 pt, conference]{ieeeconf}  

\IEEEoverridecommandlockouts                              

\usepackage{graphicx} 
\usepackage{amsmath} 
\usepackage{amssymb}  
\usepackage{amsfonts}
\usepackage{dsfont}
\usepackage{booktabs}
\usepackage{color}
\usepackage{comment}
\usepackage[dvipsnames]{xcolor}
\usepackage{algorithm}
\usepackage{algorithmic}

\let\labelindent\relax
\usepackage{enumitem}

\usepackage{xspace} 
\newcommand{\pc}{point cloud\xspace}
\newcommand{\pcs}{point clouds\xspace}
\newcommand{\ours}{FoldPath\xspace}
\newcommand{\task}{OCMG\xspace}
\newcommand{\bO}{\boldsymbol{O}}

\newcommand{\by}{\boldsymbol{y}}
\newcommand{\bp}{\boldsymbol{p}}
\newcommand{\bP}{\boldsymbol{P}}
\newcommand{\bQ}{\boldsymbol{Q}}
\newcommand{\bv}{\boldsymbol{v}}
\newcommand{\bx}{\boldsymbol{x}}
\newcommand{\bz}{\boldsymbol{z}}
\newcommand{\bw}{\boldsymbol{w}}
\newcommand{\bh}{\boldsymbol{h}}
\newcommand{\bA}{\boldsymbol{A}}

\title{\LARGE \bf
\ours: End-to-End Object-Centric\\ Motion Generation via Modulated Implicit Paths
}

\author{Paolo Rabino$^{1,2}$, Gabriele Tiboni$^{1}$, Tatiana Tommasi$^{1}$
\thanks{
This study was carried out within the FAIR - Future Artificial Intelligence Research and received funding from the European Union Next-GenerationEU (PIANO NAZIONALE DI RIPRESA E RESILIENZA (PNRR) – MISSIONE 4 COMPONENTE 2, INVESTIMENTO 1.3 – D.D. 1555 11/10/2022, PE00000013). This manuscript reflects only the authors’ views and opinions, neither the European Union nor the European Commission can be considered responsible for them. 
\newline
The EFORT group provided the authors with 
original object meshes, trajectory data, and access to the proprietary spray painting simulator used during the experiments. 
}%
\thanks{$^{1}$
Politecnico di Torino, Italy 
        {\tt\small first.last@polito.it}}%
\thanks{$^{2}$
Italian Institute of Technology, Genova}
}

\begin{document}


\definecolor{orange}{RGB}{255,229,204} 
\definecolor{lred}{RGB}{255,204,204} 
\definecolor{lgreen}{RGB}{200,240,200} 
\definecolor{lblue}{RGB}{204,229,255} 

\definecolor{darkbrown}{rgb}{0.4, 0.26, 0.13}
\definecolor{amethyst}{rgb}{0.6, 0.4, 0.8}
\definecolor{blue-violet}{rgb}{0.54, 0.17, 0.89}
\definecolor{caputmortuum}{rgb}{0.35, 0.15, 0.13}
\definecolor{darkviolet}{rgb}{0.58, 0.0, 0.83}
\definecolor{lavender}{rgb}{0.9, 0.9, 0.98}
\definecolor{indigo}{rgb}{0.29, 0.0, 0.51}

\newcommand{\tat}[1]{{\color{darkviolet}{#1}}} 
\newcommand{\paolo}[1]{{\color{ForestGreen}{#1}}} 
\newcommand{\tib}[1]{{\color{lavender}{#1}}} 
\newcommand{\todo}[1]{{\color{red}{#1}}} 
\newcommand{\rev}[1]{{\color{black}{#1}}} 
\newcommand{\placeholder}[1]{{\color{teal}#1}}

\maketitle
\thispagestyle{empty}
\pagestyle{empty}

\begin{abstract}
Object-Centric Motion Generation (OCMG) is instrumental in advancing automated manufacturing processes, particularly in domains requiring high-precision expert robotic motions, such as spray painting and welding.
To realize effective automation, robust algorithms are essential for generating extended, object-aware trajectories across intricate 3D geometries.
However, contemporary OCMG techniques are either based on ad-hoc heuristics or employ learning-based pipelines that are still reliant on sensitive post-processing steps to generate executable paths. 
We introduce \textit{\ours}, a novel, end-to-end, neural field based method for OCMG. Unlike prior deep learning approaches that predict discrete sequences of end-effector waypoints, \ours learns the robot motion as a continuous function, thus implicitly encoding smooth output paths. 
This paradigm shift eliminates the need for brittle post-processing steps that concatenate and order the predicted discrete waypoints.
Particularly, our approach demonstrates superior predictive performance compared to recently proposed learning-based methods, and attains generalization capabilities even in 
real industrial settings, where only a limited amount of 
expert samples are provided.
We validate FoldPath through comprehensive experiments in a realistic simulation environment and introduce new, rigorous metrics designed to comprehensively evaluate long-horizon robotic paths, thus advancing the OCMG task towards practical maturity.
\end{abstract}


\section{INTRODUCTION}

Automated robotic spray painting is an essential component of modern manufacturing, involving ad-hoc motion generation to replicate expert strategies while optimizing material usage, paint coverage, and engineering costs. However, planning efficient trajectories over complex 3D objects remains challenging, often requiring \emph{tens} of long-horizon paths per object, with execution times reaching 8–10 minutes.

Traditional methods for robotic spray painting predominantly rely on heuristic strategies and rule-based motion generation \cite{xi2003multiobj, Biegelbauer2005TheIA, Li2010autotraj, gleeson2022genopt}. 
These approaches typically involve partitioning 3D objects into convex regions and applying coverage planning techniques, such as raster-based paths or predefined tool orientations. While effective in controlled environments, these heuristics have limited adaptability to new object shapes, often require manual tuning, and are computationally expensive for complex surfaces.

To overcome these limitations, learning-based approaches have been recently explored to predict painting trajectories directly from expert demonstrations~\cite{tiboni2023paintnet,tiboni2025maskplanner}, given a \pc encoding of the 3D object. Such methods formulate the problem as the prediction of unordered end-effector poses, which are then post-processed into coherent paths. 
As a result, these approaches are limited to discrete path representations and yet depend on brittle post-processing steps that are highly sensitive to variations in network predictions, often leading to unfeasible trajectories and discontinuities.

\begin{figure}[t!]
    \centering
    \includegraphics[width=1\linewidth]{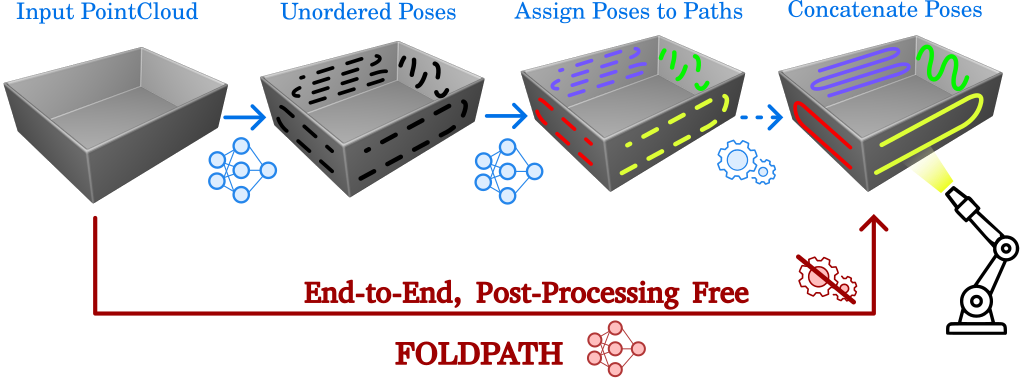}
    \caption{FoldPath tackles Object-Centric Motion Generation with an end-to-end deep learning pipeline for the first time. Differing from previous works that require multiple learning and post-processing stages, our model directly predicts long-horizon output paths that generalize across free-form 3D objects.
    }
    \label{fig:enter-label}
    \vspace{-6mm}
\end{figure}

In this work, we build on existing learning-based methods to investigate strategies that can quickly generalize motion generation for previously unseen free-form 3D objects.
Yet, we depart from multi-stage post-processing-based techniques and propose \emph{FoldPath}, the first end-to-end training pipeline for direct prediction of multiple, long-horizon paths that attain smoothness by design (see Fig. \ref{fig:enter-label}).
Particularly, we learn to decode the output paths from the input object geometry and leverage neural fields~\cite{xie2022neural} to formalize the paths as continuous functions of 6D end-effector poses. 
This allows to sample the paths by inputting scalars interpretable as relative time-steps indicating instants along the path execution.  
This paradigm makes our pose predictions ordered and smooth,
working around the shortcomings of recent approaches. 

Our data-driven model contribute to the solution of Object-Centric Motion Generation (OCMG)~\cite{tiboni2025maskplanner} tasks beyond robotic spray painting, as it does not rely on domain-specific knowledge or task-specific optimization objectives.
Importantly, we also introduce a new set of metrics for the OCMG problem, combining set-to-set and curve similarity measures. We adapt the standard AP metric~\cite{lin2014microsoft} for arbitrary-length curves, ensuring a mature evaluation of path properties like order, location and orientation precision, and uniqueness. 

In summary, our contributions are the following:
\begin{itemize}[leftmargin=*]
    \item We introduce \ours, a novel deep learning approach for generating 
    multiple smooth, long-horizon paths conditioned on 3D \pc; 
    \item We formalize a new set of evaluation metrics, bringing the task of \task to maturity for benchmark comparison within the community;
    \item We empirically find that our model achieves state-of-the-art predictive performance on the recently introduced PaintNet dataset~\cite{tiboni2025maskplanner}, including object categories with high shape variability and a limited amount of training samples. 
\end{itemize}
The results obtained in spray painting simulation confirm \ours robustness and effectiveness, indicating a high technology readiness level and its potential for real-world deployment. 
\section{Related Works}

\textbf{Spray-Painting.} Finding the optimal trajectory for paint distribution across arbitrary 3D shapes exemplifies the NP-hard coverage path planning (CPP) problem. Existing solutions mainly rely on hand-crafted heuristics designed by expert engineers, which are costly and inherently limited in generalization \cite{xi2003multiobj, Biegelbauer2005TheIA, Li2010autotraj, gleeson2022genopt}. A few works restrict the task to simplified object geometries and apply reinforcement learning \cite{kiemel2019paintrl} or propose to adapt externally provided trajectory candidates without directly handling motion generation \cite{gleeson2022genopt}. 
More recently, data-driven deep learning based methodologies demonstrated that it is possible to model expert trajectories without the need for domain-specific assumptions \cite{tiboni2023paintnet, tiboni2025maskplanner}. This paves the way to more general advancements in a set of related OCMG tasks as automated surface finishing \cite{caggiano2021automated}, and welding \cite{zhou2021welding}.
Still, the proposed methods combine multiple learning and post-processing stages that may hinder the smoothness and applicability of the predicted trajectories. 

\textbf{Robot Programming by Demonstration.}
The Programming by Demonstration (PbD)~\cite{billard2008robot} paradigm enables robots to acquire task-specific motion skills from expert demonstrations, eliminating the need for manual trajectory design. 
Early PbD methods relied on kinesthetic teaching or teleoperation, wherein an operator physically guides the robot along the desired trajectory.
Later, learning-based PbD methods were proposed to achieve generalization to new conditions or tasks, while attaining fast motion generation.
In this context, Imitation Learning (IL)~\cite{behav_cloning} methods frame PbD as a supervised learning problem where optimal actions are learned from a dataset of ground-truth data pairs, \emph{e.g.} through regression techniques~\cite{ross2011reductionbc}.
Although recent adaptations of IL investigate trajectory learning~\cite{duan2024structured}, multiple path generation~\cite{SRINIVASAN2021598}, and \pc data~\cite{Ze2024DP3diffusionpolicy}, no previous IL work copes with the OCMG problem setting---the prediction of an unknown number of output paths, of varying lengths, and uncorrelated in time~(cf. Sec. II.A in~\cite{tiboni2025maskplanner}).

\textbf{3D Deep Learning.} Deep learning models can elaborate on 3D data in various forms, \emph{e.g.}   multi-view images, depth-maps, voxel grids, meshes, or \pc. The latter are largely adopted in robotics applications thanks to their direct availability from common sensors. Dedicated network modules have been developed to manage the unordered and non-Euclidean nature of point clouds starting from point convolution \cite{qi2017pointnet, qi2017pointnet++, thomas2019kpconv} to the more recent transformers that leverage the self-attention mechanism for \pc processing \cite{zhao2021point, wu2024point}. 
Classification, segmentation, and reconstruction are among the key tasks explored in 3D deep learning. In particular, reconstruction involves processing 3D data as both input and output. 
Early approaches in \pc reconstruction utilized latent codes to produce fixed-size \pcs \cite{achlioptas2018learning}. However, these methods proved challenging to optimize and were surpassed by techniques that learn to deform point grids \cite{groueix2018papier, yang2018foldingnet}. Notably, Folding-Net \cite{yang2018foldingnet} proposed to iteratively deform a fixed 2D grid, lifting it into higher dimensions until the desired shape is achieved. More recently, combinations of multiple small Folding-Nets and transformers demonstrated highly accurate \pc reconstructions \cite{ Yu2023AdaPoinTrDP, Duan2024TCorresNetTG}.

\textbf{Neural Fields.} 
3D signals can also be represented as continuous functions parametrized by neural networks. Neural fields are such networks that elaborate on discrete coordinate-based data and implicitly encode the underlying shape function, which can then be queried at any point \cite{xie2022neural}.
These models have found applications in diverse areas, including scene reconstruction \cite{mildenhall2020nerf}, grasping \cite{karunratanakul2020grasping}, and physics simulation \cite{Cleach2022DifferentiablePS, irshad2024neuralfieldsroboticssurvey}. While deep Multilayer Perceptrons (MLPs) are commonly employed as neural fields, the incorporation of Fourier features alongside coordinate inputs has demonstrated improved performance for high-frequency signals \cite{tancik2020fourier}. This insight has led to the development of architectures utilizing sine \cite{sitzmann2019siren} or other periodic activations \cite{liu2024finer}. 
Neural fields have traditionally represented individual instances, with each shape or scene requiring its own parameterized model.
To address this limitation, research has explored conditioning neural fields on latent variables to encode multiple fields. Early approaches involved simple feature concatenation \cite{park2019deepsdf}, while subsequent work introduced modulation techniques \cite{mehta2021modulated}.
\section{Method}
Given a 3D \pc as input, our goal is to generate the paths a spray painting robot should execute to complete the painting task. Since the number and length of paths vary significantly depending on the object geometry, a flexible and adaptive solution is required. 

To address this, we propose a novel deep learning model that encodes the object’s \pc into a compact representation, which is then decoded into a predefined set of path embeddings. These embeddings act as codewords for dedicated path heads, guiding the generation of points in a 6D vector field defining the spray gun nozzle position and orientation using a conditioned neural field. Each codeword is leveraged multiple times to produce an ordered set of poses that collectively form a path. Additionally, the model predicts a confidence score for each path, enabling the selection of the most relevant ones. 

The intuition at the basis of our approach is that the obtained codewords implicitly contain all the information needed to define a continuous path curve, and each path head operates as a \emph{sampler} and \emph{neural field generator} conditioned by the codeword.

\begin{figure*}
    \centering
    \includegraphics[width=0.95\linewidth]{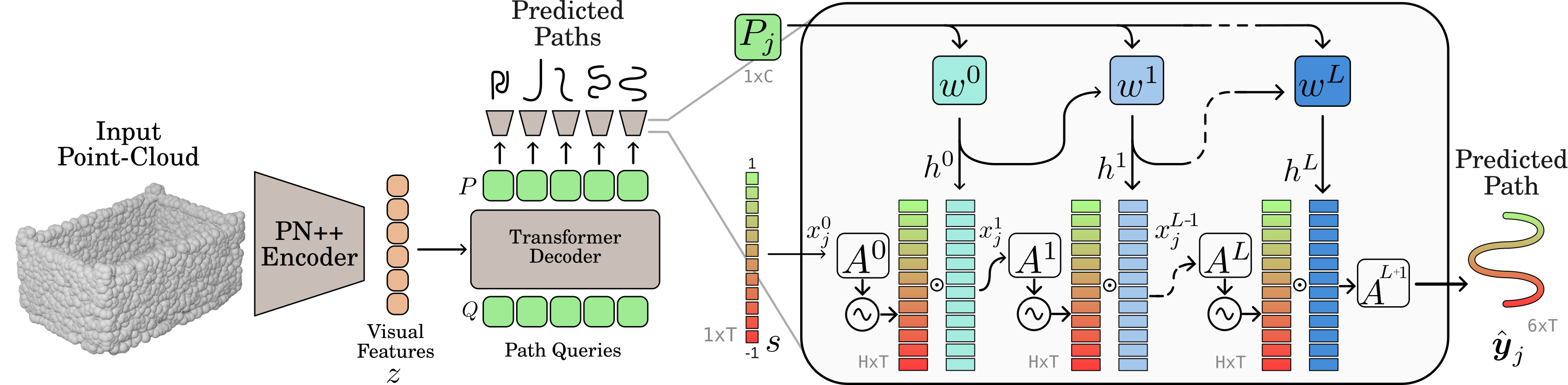}
    \caption{
    Schematic visualization of \ours. It first encodes the input \pc using a PointNet++ backbone  \cite{qi2017pointnet++}, then it elaborates the obtained visual features $\bz$ with a transformer decoder together with the learned queries $\bQ_j$ to obtain path embeddings $\bP_j$. Finally, every path-specific neural field inspired head is fed with $\bP_j$ and a scalar $x_{j,t}$ to yield the $t$-th 6D pose $\hat{\by}_{j,t}$ of the output path. The whole path is generated by sampling $T$ scalars $s_{t=1,\ldots,T} \in [-1,1]$.
    }
    \label{fig:model}
    \vspace{-4mm}
\end{figure*}

The sampling is guided by the user who decides on the \emph{arbitrary number of 6D points} needed to quantize and execute the path, while the network progressively \emph{fold} the curve composed by these points and their embeddings to fit the ground truth path. 
Remarkably, by feeding the network with a list of relative curve locations (interpretable as time-steps along the path execution), the output will seamlessly describe the corresponding path through its ordered components, avoiding the post-processing steps needed in previous approaches.  By modeling the path head with MLP layers, we leverage their biases to generate predictions that are already smooth and ordered by design.
Our approach significantly differs from current LLM-based planners, such as those described by \cite{kim24openvla}, which are limited to predicting a single path in discrete steps and necessitate billions of parameters. In contrast, our simpler architecture can predict a variable number of complete paths, each with an adaptable number of points.
We name our method \ours and present its schematic visualization in Fig. \ref{fig:model}. In the following we formally define the model together with its optimization objective (Sec. \ref{sec:overview}), and describe the adopted architecture (Sec. \ref{sec:architecture}). Finally, we provide details on the waypoint sampling (Sec. \ref{sec:wsampling}). 
\subsection{\ours Overview}
\label{sec:overview}
Let $\bO$ represent the object geometry as a \pc consisting of an arbitrary number of points in 3D space. Each object $\bO$ is associated with a ground truth set of paths $\{\by_i \}_{i=1}^{n(\bO)}$, with object-dependent cardinality $n(\bO)$. 
\ours employs an encoder to map $\bO$ to the visual features 
$\bz \in \mathbb{R}^{256\times C}$. A transformer decoder then processes $\bz$ alongside $N > n(\bO)$ path query vectors $\bQ_{j=1,\ldots,N} \in \mathbb{R}^C$ and outputs $\bP_{j=1,\ldots,N} \in \mathbb{R}^C$, which define specific path prototypes. 
Each of them is further elaborated by a tailored network head, which also takes as input a scalar value $x_j$ and outputs one of the 6D points of the path $\hat{\by}_j$. By considering $T$ scalars $s_{t=1,\ldots,T} \in [-1,1]$ and assigning $x_{j,t}=s_t$, the output set $\hat{\by}_{j,t=1,\ldots,T}$ will represent 6D points sampled from the predicted $j$-th path. Specifically, we define $\hat{\by}_{j,t}=\{\hat{\bp}_{j,t},\hat{\bv}_{j,t}\}$ with the two sub-vectors respectively representing 3D position and orientation of each 6D point. 

Considering the difference in cardinality between the predicted paths and ground truth ones, a matching procedure is needed. The ground truth paths are sampled by using the same set $s_{t=1,\ldots,T}$ adopted for the predictions. Here the scalars are interpreted as relative positions along the path: $-1$ corresponds to the first point, $0$ to the middle, and $1$ as the last one, with other intermediate points derived via linear interpolation. Thus, we obtain the set of reference ground truth points $\by_{i,t=1,\ldots,T}$ and assign a confidence score $f_{i=1,\ldots, n(\bO)}=1$ to all the corresponding paths. 

To perform a bipartite matching, we add extra $N-n(\bO)$ zero-padded paths with score $f_{i=n(\bO)+1,\ldots,N}=0$ and we search among all permutations of N elements $\mathfrak{S}_N$ via the Hungarian algorithm by relying on the 3D point locations: 
\begin{equation}
    \hat{\sigma} = {\arg \min}_{\sigma \in \mathfrak{S}_N} \sum_i^N\mathcal{L}_{\text{match}}(\by_{\sigma(i)}, \hat{\by}_i)~, 
\end{equation}
\begin{equation}
    \text{where} \quad \mathcal{L}_{match} = \mathds{1}_{f_i \ne 0} (||\bp_{\sigma(i)} - \hat{\bp}_i||_2)~.
\end{equation}

Finally, the overall objective function of \ours is $\mathcal{L}=\mathcal{L}_{points} + \mathcal{L}_{conf}$ defined by leveraging the optimal matching $\hat{\sigma}$ and the predicted confidence score $ \hat{f}_i$ with  
\begin{equation}
    \resizebox{0.9\hsize}{!}{$
    \mathcal{L}_{points} = \sum_i^N \mathds{1}_{f_i\ne 0} \left\{||\bp_{\hat{\sigma}(i)} - \hat{\bp}_i||_2 + 1- \frac{\bv_{\hat{\sigma}(i)}\cdot \hat{\bv}_i}{||\bv_{\hat{\sigma}(i)}||_2\cdot||\hat{\bv}_i||_2} \right\}$     }
\end{equation}
\begin{equation}
    \mathcal{L}_{conf}=\sum_i^N -(1-\hat{f}_i)^\gamma\log \hat{f}_i~.
    \vspace{-2mm}
\end{equation}

In short, the \emph{points} loss evaluates the 6D points set similarity by combining the Euclidean distance between their 3D location and the angular (cosine) distance between the corresponding orientation vectors. The \emph{confidence} prediction correctness is measured by the focal loss \cite{lin2017focal} with $\gamma=2$. 

\subsection{Architecture}
\label{sec:architecture}

We design \ours to process 3D \pcs of objects and learn an implicit representation of path curves, enabling arbitrary sampling of the generated paths. 
An initial encoder downsamples the \pc by using point-convolution layers and then upsamples it via feature propagation layers to output $\bz$
that we interpret as visual features. A transformer decoder operates on $\bz$, and on the path queries $\bQ_j$ by alternating layers of self-attention between the path queries and cross-attention between the path queries and visual features to generate the path propotypes $\bP_j$. The queries $\bQ_j$ are randomly initialized and then learned during training.

Each path prediction head takes as input a single path prototype $\bP_j$ and elaborates the initial scalar value 
which represents the progress along the path 
via $L$ stacked modulated~\cite{mehta2021modulated} MLP blocks.
So, starting from $\bx_{j,t}^{l=0}=x_{j,t}$ we 
obtain $\bx_{j,t}^{l=1},\ldots,\bx_{j,t}^{l=L}$. More precisely, each block receives an embedding $\bx_{j,t}^l$ and creates the next one $\bx_{j,t}^{l+1}$ as follows
\begin{equation}
    \bx_{j,t}^{l+1} = \bh_j^l\odot act(\bA^l\cdot\bx_{j,t}^l)~, 
\end{equation} 
where $l$ is the block index,
$act$ is a non-linear activation function that will be discussed later,  and $\bA^l$ is a linear layer with dimension $1\times H$ for $l=0$ and $H \times H$ otherwise. 
Specifically, $\bh_j^l$ is the hidden representation that modulates the $l$-th MLP block depending on the input path prototype $\bP_j$, defined by 
\begin{equation}
\resizebox{0.9\linewidth}{!}{
    $\bh_j^0 = \text{ReLU}(\bw^0 \bP_j), \quad \bh^{l+1}_j = \text{ReLU}(\bw^{l+1}\left[h^l_j \bP_j\right]^\top).$
    }
\end{equation}
Here $\bw^0$ has dimension $H \times C$, while for $l\neq 0$, $\bw^l$ has dimension $H \times (H+C)$. The last MLP block outputs $\bx_{j,t}^L$ which is then fed to a final linear layer $\bA^{L+1}$ that maps it to the 6D prediction $\hat{\by}_{j,t}$.

Regarding the choice of $act$, we follow the neural field literature, which also inspired our use of modulated blocks for continuous visual signal representation.
Previous work \cite{sitzmann2019siren,mehta2021modulated,liu2024finer} demonstrated that activation functions belonging to the sinusoidal family are particularly suitable when dealing with signals with high-frequency components. Thus, we adopted siren activations \cite{sitzmann2019siren} and finer activations \cite{liu2024finer} in addition to the standard ReLU. 

Finally, the path head further elaborates on $\bP_j$ via a two-layer Feed Forward Network ending with a sigmoid activation that outputs the confidence score $\hat{f}_j$. 

\subsection {Waypoint Sampling} 
\label{sec:wsampling}
We described the model flow by considering the path quantized in $t=1,\ldots,T$ waypoint components. Still, we highlight that the network can elaborate a whole point sequence in parallel without any downside other than memory consumption. Indeed, to let the model reason on path curves during training, we interpolate the ground truth path points to then re-sample the underlying curve at an arbitrary rate. We adopt a uniform random sampling in training as it facilitates the learning dynamics (see Sec. \ref{sec:ablation}) and at test time we use an equidistant point list by setting $s_t=-1+t\frac{2}{T}$. 

\section{Metric}
\label{sec:metric}

Spray painting paths are inherently continuous curves. Measuring the performance of a path prediction model that outputs discrete sequences of 6D poses requires applying an analogous sampling on the ground truth paths and the adoption of set-to-set metrics as the point-wise chamfer distance (PCD) \cite{guo2020deep}. 
This is the evaluation strategy used in previous work \cite{tiboni2023paintnet}\cite{tiboni2025maskplanner} that, however, depends on the sampling rate, disregards the point order as well as the identity of each path, and focuses on a preliminary stage before the refinement needed to determine the spray plan for execution. Moreover, the PCD exhibits a critical sensitivity to outliers, with even a few inaccurate points leading to a skewed PCD mean. 
Thus, the assessment may become unreliable, particularly in scenarios requiring fine-grained comparisons. 

To address these limitations, we propose a novel Average Precision (AP) metric based on the F-score and leveraging the Dynamic Time Warping (DTW) distance \cite{1163055}. 
Calculating the F-Score is a commonly used strategy in \pc prediction tasks \cite{topnet2019, Yu2023AdaPoinTrDP} with the chamfer distance used to obtain the point matching, but we adopt the DTW to ensure the correct point ordering. 

More formally, given the ground truth and the predicted 3D point lists $\boldsymbol{p} \in \mathbb{R}^{K \times 3}$ and $ \boldsymbol{\hat p}\in \mathbb{R}^{M \times 3}$,
we find the optimal match between each component point such that the DTW distance is minimized. 
This distance encodes both the point locations and their index order in the list. 
In this way, similar paths with low distances will have the same shape and point sorting, while the metric remains agnostic to different sampling rates as each point from one set can match multiple points from the other and vice versa.
For simplicity, we use the wildcard $*$ to indicate both the DTW ground truth match for a predicted point (precision) and the DTW predicted point match for a ground truth point (recall). 
So when comparing the two 6D path poses corresponding to the 3D point lists, we have F-score$=2(pre \times rec)/(pre + rec)$, where 
\begin{align}
rec &= \frac{1}{K}\sum_{k=1}^K \mathds{1}(||\boldsymbol{p}_k-\hat {\boldsymbol{p}}_{*}||_2 < \delta, \measuredangle(\bv_k,\hat{\bv}_*)<\theta)  \\
pre &= \frac{1}{M}\sum_{m=1}^M \mathds{1}(||\hat {\boldsymbol{p}}_m- \boldsymbol{p}_{*}||_2 < \delta, \measuredangle(\hat{\bv}_m,\bv_*)<\theta)~.  
\end{align} 
Here we use two thresholds for the Euclidean and angular distances with $\delta$ set to $0.025$ in normalized space, and $\theta=10^\circ$, similarly to \cite{Yu2023AdaPoinTrDP}.
In summary, the F-score provides values in $[0, 1]$ 
that express the similarity between two paths of arbitrary length. 
By using this score, we can calculate the precision-recall curve and compute the AP, 
thus evaluating the capability of any path prediction model.
For the spray painting task, path execution does not have a preferential direction as long as the points along a path are correctly ordered. Hence, we calculate the F-score considering both the order with which the path points are generated and its reverse, keeping the highest value. 
For the experimental evaluation, we report two AP metrics. $AP_{DTW}^{50}$ is obtained by considering that a path is correct when the F-Score is at least 0.5, and represents the case when prediction and ground truth share at least 50\% of the points. $AP_{DTW}$ is the mean of APs calculated with a threshold ranging from 0.5 to 0.95 with 0.05 step increments: this represents a comprehensive metric accounting for the overall quality of the predictions.
\section{Experiments}
\label{sec:experiments}
In our experimental analysis, we assess the capabilities of our method in predicting paths that are ready for direct execution on a robot. Indeed, \ours does not require a post-processing phase on the output paths, which is instead essential for existing approaches. We consider the spray painting task as a relevant industrial use case and thoroughly evaluate the performance of \ours, further conducting ablation studies to gain deeper insights into its inner workings. 

\subsection{Dataset and Baselines} 
We align with previous work \cite{tiboni2023paintnet, tiboni2025maskplanner} and adopt the PaintNet dataset~\cite{tiboni2025maskplanner} for our experimental benchmark, covering four object categories of increasing complexity: cuboids, windows, shelves, and containers.
Each data sample includes the input object as a 3D \pc of 5120 points, and the set of associated ground-truth paths encoded as a discrete sequence of 6D waypoints. More precisely, 6D poses describe the ideal paint deposition location in $(x,y,z)$ coordinates, and the end-effector orientation of the gun nozzle as a 3D unit vector (2-DoF representation).

We consider three reference baselines introduced in \cite{tiboni2025maskplanner}, briefly described below for clarity. 
\emph{Path-wise} directly outputs a set of paths, as fixed sequences of 6D end-effector poses. For each pose and each path, a confidence score is simultaneously predicted. As a result, this model attempts to autonomously retain the correct number and lengths of output paths in end-to-end fashion, similarly to our approach.
\emph{Autoregressive} starts by predicting a set of initial configurations for each output path. Then, each path is independently generated in an autoregressive manner, by predicting 6D poses and a termination probability.
\emph{MaskPlanner}~\cite{tiboni2025maskplanner} predicts a set of path segments, \emph{i.e.} short sequences of 6D poses.
Its key strength lies in generating path segments as local sub-goals for the robot to follow. However, MaskPlanner relies on a final post-processing step for sub-goal concatenation, which may ultimately lead to unfeasible paths.
\subsection{Implementation Details} 
We train \ours separately for each object category for 200 epochs 
with a batch size of 24. We use the Adam optimizer \cite{Kingma2014AdamAM} and cosine annealing scheduling (with learning rate from 3e-4 to 1e-8). 
The encoder is a PointNet++ backbone \cite{qi2017pointnet++},  the same used in \cite{tiboni2023paintnet, tiboni2025maskplanner},   
with the last feature propagation layer removed. The encoder is trained from scratch.
The dimension of the visual features $\bz$ is $256\times C$ with $C=384$. 
The transformer decoder has 4 layers with 4 heads, an hidden dimension of $C$ and takes as input $N=40$ learned path queries. The path head has $L=4$ layers and an hidden dimension of $H$=512. The waypoint cardinality is set to $T$=64 in training and $T$=384 at testing.  

We implement three different versions of our model: \ours (\textit{ReLU}), \ours (\textit{Siren}), and \ours (\textit{Finer}). Inspired by the blocks in the decoder of FoldingNet \cite{yang2018foldingnet}, every layer in our path head is composed of linear, ReLU, and layer norm. The Siren and Finer versions include respectively sine activation \cite{sitzmann2019siren} and finer activation \cite{liu2024finer} instead of ReLU and don't use normalization due to the constrained nature of the activation functions.

\subsection{Results on Cuboids, Windows, Shelves}
\newcommand{\oursR}{\ours (\textit{ReLU})}
\newcommand{\oursS}{\ours (\textit{Siren} \cite{sitzmann2019siren})}
\newcommand{\oursF}{\ours (\textit{Finer} \cite{liu2024finer})}

\begin{table}[t!]
     \caption{Comparison between baselines and \ours on three categories of the PaintNet benchmark \cite{tiboni2025maskplanner}. MaskPlanner shows top PCD but falls off when post-processing is applied. The full path AP indicates superior results for \ours. 
    }
    \centering
    \resizebox{0.9\linewidth}{!}{
    \begin{tabular}{l| c c c}
         \toprule
         \multicolumn{4}{c}{\textbf{CUBOIDS}}\\
         \midrule
         Model & $PCD \downarrow$  & $AP_\text{DTW}^{50} \uparrow$ & $AP_\text{DTW} \uparrow$ \\
         \midrule
         Autoregressive  & 33.2     & 2.1    & 0.0   \\
         Path-Wise   & 48.0     & 32.9   & 10.8  \\
         MaskPlanner (W/O Post) & 6.5 & -  & -  \\
         MaskPlanner (W Post)  & 19.9  &  80.0 & 50.3\\
         \oursR      & 11.2     & 59.8   & 35.2  \\
         \oursS      & 9.2     & 97.5   & 60.3  \\
         \oursF      & \textbf{5.5}     & \textbf{99.2}   & \textbf{91.1}  \\
         
         \toprule
         \multicolumn{4}{c}{\textbf{WINDOWS}}\\
         \midrule
          Model & $PCD \downarrow$  & $AP_\text{DTW}^{50} \uparrow$  & $AP_\text{DTW} \uparrow$ \\
         \midrule
         Autoregressive & 64.1 & 10.5 & 4.0 \\
         Path-Wise & 45.5  & 28.0 & 11.1\\
         MaskPlanner (W/O Post) & \textbf{6.8}  & - & - \\
         MaskPlanner (W Post) & 336.0  & 70.4 & 50.4 \\
         \oursR &  106.7 & 91.4 &  71.8 \\
         \oursS &  114.3 & 91.3 &  71.9 \\
         \oursF &  104.6  & \textbf{91.9} &  \textbf{75.0}\\
         \toprule
         \multicolumn{4}{c}{\textbf{SHELVES}}\\
         \midrule
         Model & $PCD \downarrow$ & $AP_\text{DTW}^{50} \uparrow$ & $AP_\text{DTW} \uparrow$ \\
         \midrule
         Autoregressive  & 40.45 & 23.6 & 8.5  \\
         Path-Wise & 46.7 & 9.7 & 2.5 \\
         MaskPlanner (W/O Post) & \textbf{7.4} & - & - \\
         MaskPlanner (W Post) & 1470 & 35.1 & 26.0 \\
         \oursR & 65.8 & 88.4 &  75.4 \\
         \oursS & 64.7 & 89.5 &  78.0 \\
         \oursF & 65.9 & \textbf{91.3} &  \textbf{84.3} \\
         \bottomrule
    \end{tabular}
    }
    \vspace{-16pt}
    \label{tab:main}
\end{table}
We run a first set of experiments on the three object categories with synthetic spray painting paths: cuboids, windows, and shelves with the same train/test splits of \cite{tiboni2025maskplanner}. For MaskPlanner we consider both the final prediction after post-processing (W Post) and its raw output without post-processing (W/O Post). 
The latter consists of short sequences of 6D points that are grouped as part of specific paths but remain unconnected. As a result, they cannot be directly evaluated as paths using the AP metric.

The results 
in Tab. \ref{tab:main} indicate that for the cuboids MaskPlanner provides much lower PCD 
than the Autoregressive and Path-Wise baselines. In terms of AP, MaskPlanner (W Post) outperforms the competitor baselines and also 
\ours with ReLU activations.  
Still, when passing to the Siren version, 
\ours improves and gets to top performance when adopting Finer activations. High $AP^{50}_{DTW}$ values indicate a high number of correct paths, while high $AP_{DTW}$ means that the paths generated by the model are overall accurate.  

For windows and shelves the AP results show a trend similar to that observed on cuboids, but the best PCD is that provided by MaskPlanner (W/O Post). On one hand, filtering and concatenating the predicted pose segments during post-processing significantly affect the PCD, and on the other provide paths with better AP than the baseline competitors (Autoregressive and Path-Wise), but largely worse than \ours in all its versions. 

In general, we can state that the AP metric reflects a model's ability to generate coherent and smooth paths---aspects on which \ours excels. 
The Autoregressive and Path-Wise baselines generate entire paths directly, but their performance is limited by optimization challenges: Path-Wise struggles with the complexity of high-dimensional outputs, whereas Autoregressive faces difficulties in making accurate long-horizon predictions, particularly for cuboids. 

Overall, the optimization challenges are also evident from the training time needed by these methods as well as by MaskPlanner, as they all require six times the number of epochs to converge in respect to \ours.

\begin{figure}
    \centering
    \includegraphics[width=0.97\linewidth]{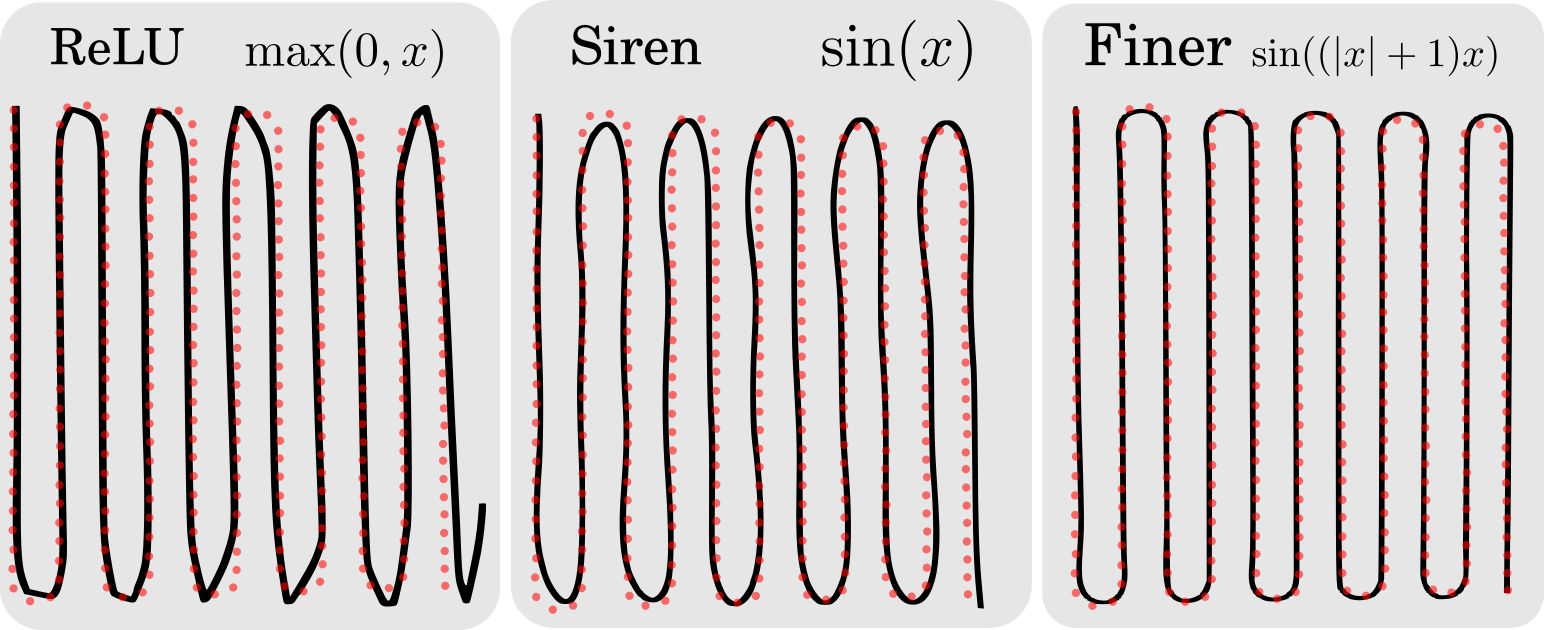} \vspace{-2mm}
    \caption{Qualitative comparison of a single path from the cuboids test set between various activation functions for our \ours model. In black the predictions, in light red the ground truth. ReLU activation yields sharp corners, undesirable for robotic applications. Siren activation yields good curves but lacks precision. Finer performs best overall.}
    \label{fig:qualitative_activation}
    \vspace{-4mm}
\end{figure}

Regarding \ours, we can further discuss the role of the chosen activation functions. 
ReLU achieves top results on tasks with a prevalence of straight paths (windows and shelves) but fails when the paths are highly complex and require multiple curves (cuboids). 
The Siren activation performs well overall but tends to favor wavy paths, leading to reduced precision in straight path sections. 
Finer activations combine both characteristics and perform well on all object categories, confirming the suitability of sinusoidal activations for high-frequency predictions (see Fig. \ref{fig:qualitative_activation}).  

\begin{figure*}
    \centering
    \includegraphics[width=0.9\linewidth]{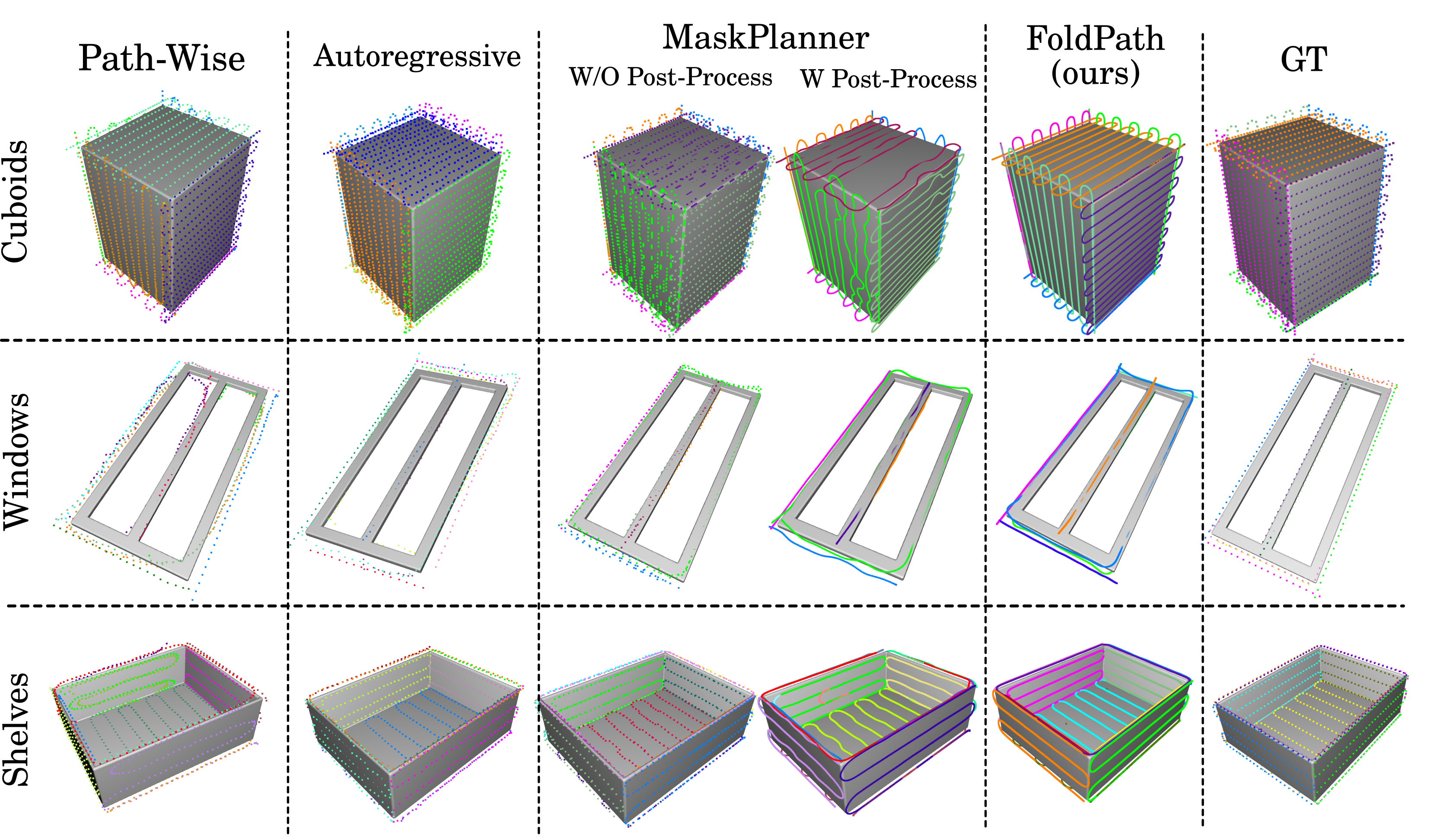}
    \caption{Qualitative results on randomly chosen samples from the three main categories of the PaintNet benchmark. Path-wise struggles with long paths. Autoregressive paths are coherent but may miss details. MaskPlanner is able to generate highly precise paths but the post-processing plays a key role in producing accurate results.  FoldPath overall outputs the clearest paths without need of any post-processing step.}
    \label{fig:qualitative}
\end{figure*}

In Fig. \ref{fig:qualitative}, we present qualitative results for all baselines and the best version of \ours. Path-Wise exhibits unprecise paths that deviate significantly from the ground truth. Autoregressive predictions are qualitatively good but are subtly imprecise, confirming what already observed from the quantitative results.  
MaskPlanner benefits significantly from post-processing, but small errors in the original prediction cause significant deviations in the final paths. \ours generates smooth and precise paths, thanks to its ability to implicitly encode the path curve rather than restricting its focus to discrete poses. 

\subsection{Results on Containers}
Here, we zoom in on the challenging container class of the PaintNet dataset, which is limited to 88 samples  (70 train, 18 test) collected from a real industrial scenario.
Container instances particularly show high shape variability (\emph{e.g.}, wavy and grated surfaces), and are paired with spray painting paths that are irregular due to the recording from real-world manual executions. To fairly evaluate the performance of the path generative models on this testbed, we adopt a version of the AP metric that is more lenient to errors. We define 
$AP_{DTW}^{easy}$ that sweeps the DTW threshold from 0.05 to 0.5 instead of 0.5 to 1.0. This change maintains all the desired metric properties described in Sec. \ref{sec:metric}, but with a lower number of points needed to match between the ground truth path and the model output to be considered as a correct prediction.
Following the protocol defined in \cite{tiboni2023paintnet}, we also use a proprietary simulator developed by EFORT to test the actual ability of \ours to paint the real objects. We report the paint coverage (as defined in~\cite{tiboni2025maskplanner}) in percentage points averaged across all test instances. 

\begin{figure}
    \centering
    \includegraphics[width=0.9\linewidth]{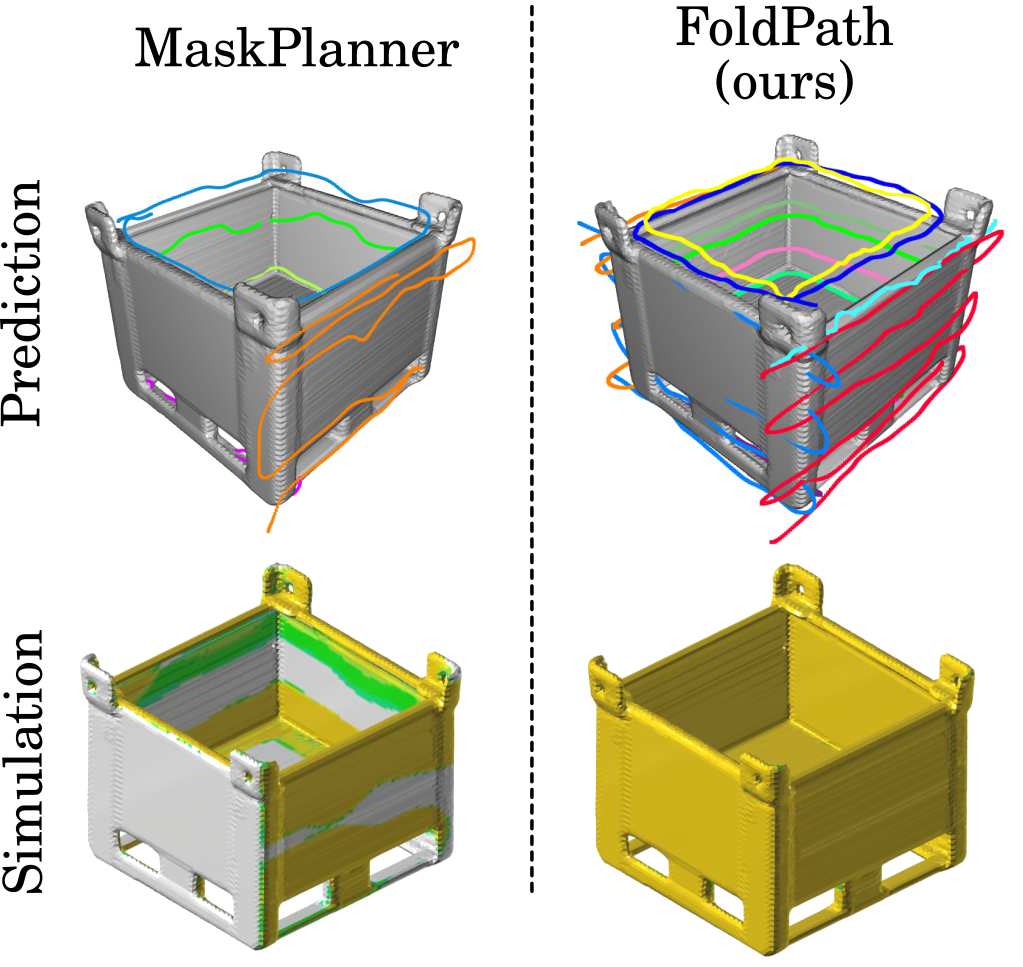}
    \caption{Qualitative evaluation in simulation for a random sample from the container test set.
    The colormap ranges from green (low) to yellow (high) paint deposition.
    Some paths can be hard to see due to compenetration with object mesh, but in practice the nozzle is 12 cm far from the object so even compenetrating paths yield correct painting executions.}
    \label{fig:container-qualitative}
    \vspace{-5mm}
\end{figure}

We assess our model's performance by considering its version with finer activations and training for 200 epochs, a sixth of the baseline's training. 
Tab. \ref{tab:containers} reveals that, despite the limitations highlighted by the PCD results, our method achieves improved $AP_{DTW}^{easy}$ scores compared to baselines. The obtained paint coverage percentage further confirms the effectiveness of \ours. 
The inverse trend between the paint coverage and the PCD results confirms that the latter metric is not really informative on the model's effectiveness for the spray painting task. 
The qualitative results in Fig. \ref{fig:container-qualitative} indicate that, even if the generated paths do not fully match expert-defined optimal paths, the predictions of \ours capture essential path qualities for practical applications. 

\begin{table}[t!]
    \caption{Results for the challenging container class in PaintNet \cite{tiboni2025maskplanner}. 
    MaskPlanner obtains the best PCD but it is negatively affected by post-processing. \ours shows top AC results that correlate with the effectiveness in paint coverage.
    }
    \centering
    \resizebox{0.95\linewidth}{!}{
    \begin{tabular}{l| c c c c}
         \toprule
         Model & $PCD \downarrow$  & $AP_\text{DTW}^{easy} \uparrow$ & Paint Cov. \% $\uparrow$ \\
         \midrule
         Path-Wise   & 556.0     & 9.5 &  58.4    \\
         Autoregressive  & 708.7     & 9.2  &  53.3 \\
         MaskPlanner (W/O Post) & \textbf{220.7}
         & -  & 90.9\\
         MaskPlanner (W Post) & 1483.0   &   9.6 &  29.5 \\
         \ours       &  584.1  & \textbf{13.7} & \textbf{91.1}\\
         \toprule
    \end{tabular}
    }
    \label{tab:containers}
    \vspace{-8pt}
\end{table}

\subsection{Ablation Studies}
\label{sec:ablation}
\begin{table}[t]
    \caption{$AP_{DTW}$ on the PaintNet benchmark for different design choices of \ours. In bold the adopted configuration. 
    }
    \centering
    \resizebox{0.95\linewidth}{!}{
    \begin{tabular}{l| c c c}
         \toprule
         \multicolumn{4}{c}{\textbf{Training Sampling Strategy}} \\
         \midrule
         Sampling Strategy  & CUBOIDS & WINDOWS & SHELVES \\
         \midrule
         Equispaced & 88.0 & 74.6 & 85.3 \\
         Noisy Equispaced & 90.9 & 75.4 & 86.1 \\
         \textbf{Uniform Noise} & 91.1 & 75.0 & 84.3 \\
         \midrule
         \multicolumn{4}{c}{\textbf{Decoder Size}} \\
         \midrule
         Hid. Dim / Num. Layers & CUBOIDS & WINDOWS & SHELVES \\
         \midrule
         Tiny (256/2) & 86.6 & 58.6 & 79.7 \\
         Small (256/4) & 89.0 & 60.2 & 83.9 \\
         \textbf{Medium} (512/4) & 91.1 & 75.0 & 84.3\\
         Large (512/6) & 91.4 & 66.5 & 85.6 \\         
         \midrule
         \multicolumn{4}{c}{\textbf{Conditioning Strategy}} \\
         \midrule
         Strategy & CUBOIDS & WINDOWS & SHELVES \\
         \midrule
         FoldingNet Concat. \cite{yang2018foldingnet} & 14.4 & 46.6 & 56.4 \\ 
         \textbf{Modulation} \cite{mehta2021modulated} & 91.1 & 75.0 & 84.3 \\
         \bottomrule
    \end{tabular}
    }
    \label{tab:ablations}
    \vspace{-4mm}
\end{table}

The design of \ours requires several model choices that can impact the final results. We already discussed the role of the activation functions, we focus here on the waypoint sampling strategy, on the size of the path head and the technique used to condition the path head with the path codewords obtained from the transformer decoder. In Tab.~\ref{tab:ablations} we present the results obtained while investigating these aspects. We focus on cuboids, windows, and shelves where we have a sizable amount of samples in training that allows to properly assess the model choices. Moreover, we consider only the AP metric that better reflects the accuracy of the predicted paths over the whole test set. 

When selecting the $T$ scalars $s_{t=1,\ldots,T} \in [-1,1]$ and assigning $x_{j,t}=s_t$ at training time, the exact selection procedure has minimal effect on the model's outcome. This is the conclusion we can draw from the top part of Tab.~\ref{tab:ablations}, where we consider simple equally-spaced points (corresponding to the same strategy used for sampling at test time), a uniform random selection, and an intermediate case where random Gaussian noise is added to the equispaced values. As there isn't a clearly superior single strategy, all the choices are equally valid and we adopt  the uniform noise everywhere. 

Regarding the size of the path head, we experiment with different values of the hidden dimension $H$ and the number of layers $L$. The results in the middle part of Tab. \ref{tab:ablations} highlight that a small network module has limited capacity, while a large one shows diminishing returns or may under-perform due to high optimization complexity. This supports the choice of the Medium setting for our experiments. 

Finally, we examine the role of the modulation strategy inspired by \cite{mehta2021modulated} that we use to condition the path heads with the path codewords. It differs from the solution adopted in FoldingNet \cite{yang2018foldingnet}, according to which the path codeword $\bP_j$ is simply concatenated to $\bx_j$ as input to every layer. From the bottom part of Tab, \ref{tab:ablations} it is evident that the modulation is much more effective than the concatenation solution which suffers especially when dealing with long paths as for the cuboids. The reason lies in the heavy dimensionality reduction applied on the codewords to maintain a manageable dimensionality of the concatenated features, consequently limiting the expressive abilities of the path head.    

\section{Conclusions}
In this work we propose \ours, a novel deep learning model for end-to-end Object-Centric Motion Generation from 3D \pcs.
Inspired by concepts from the neural field literature, our approach successfully predicts a large number of smooth, long-horizon paths in a single forward pass, without the need for costly heuristics or sensitive post-processing steps.
Moreover, we introduce tailored evaluation metrics for OCMG that provide useful insights into the real functioning of the model and support fair benchmark comparisons. Extensive experimental evaluations demonstrated the effectiveness of \ours for the spray painting task. 

We believe this contribution is highly relevant to the broader OCMG problem and plan to extend our findings to other tasks that feature geometrically complex paths, 
such as robotic welding and automated multi-UAV visual inspection.

\addtolength{\textheight}{-12cm}   




%



\bibliographystyle{IEEEtran}
\bibliography{IEEEbiblio}

\end{document}